\documentclass[10pt,twocolumn,letterpaper]{article}

\usepackage[pagenumbers]{cvpr}

\usepackage[T1]{fontenc}

\usepackage{graphicx}
\usepackage{amsmath}
\usepackage{amssymb}
\usepackage{booktabs}
\usepackage{tabularx}
\usepackage{multirow}
\usepackage{colortbl}
\usepackage{microtype}

\definecolor{rankfirst}{HTML}{DCEFD4}
\definecolor{ranksecond}{HTML}{FFF2CC}
\definecolor{rankthird}{HTML}{D9EAF3}

\definecolor{cvprblue}{rgb}{0.21,0.49,0.74}
\usepackage[
    pagebackref,
    breaklinks,
    colorlinks,
    allcolors=cvprblue
]{hyperref}

\def\confName{CVPR}
\def\confYear{2027}

\title{
    H3Remover: Reinforcing Video Object Removal without Paired Data
}

\author{%
    {\normalfont\fontsize{10}{12}\selectfont
        Kangjie Chen\textsuperscript{1}\thanks{Equal contribution.}\hspace{0.8em}%
        Xiangyu Li\textsuperscript{1}\footnotemark[1]\hspace{0.8em}%
        Jinhao Deng\textsuperscript{1}\hspace{0.8em}%
        Dongbin Zhang\textsuperscript{1}\hspace{0.8em}%
        Chaoda Zheng\textsuperscript{1}\hspace{0.8em}%
        Shijia Chen\textsuperscript{1}\hspace{0.8em}%
        Yuqi Zhang\textsuperscript{1}%
    }\\[0.15em]
    {\normalfont\fontsize{10}{12}\selectfont
        Choo Sin Wai\textsuperscript{2}\hspace{0.8em}%
        Boyi Sun\textsuperscript{3}\hspace{0.8em}%
        Puyi Wang\textsuperscript{4}\hspace{0.8em}%
        Yu Zhang\textsuperscript{1}\hspace{0.8em}%
        Xianming Liu\textsuperscript{1}\hspace{0.8em}%
        Boyang Wang\textsuperscript{1}\thanks{Corresponding author.}%
    }\\[0.40em]
    {\normalfont\fontsize{9}{11}\selectfont
        \textsuperscript{1}XPENG
        \hspace{1.5em}%
        \textsuperscript{2}THU
        \hspace{1.5em}%
        \textsuperscript{3}UCAS
        \hspace{1.5em}%
        \textsuperscript{4}CUHK
    }
    \\[0.35em]
    {\normalfont\fontsize{9}{11}\selectfont
    \url{https://chenkangjie1123.github.io/H3Remover/}
    }
}

\begin{document}

\makeatletter
\g@addto@macro\@maketitle{%
    \par\noindent
    \begin{minipage}{\textwidth}
        \centering
        \includegraphics[width=1.0\linewidth]
            {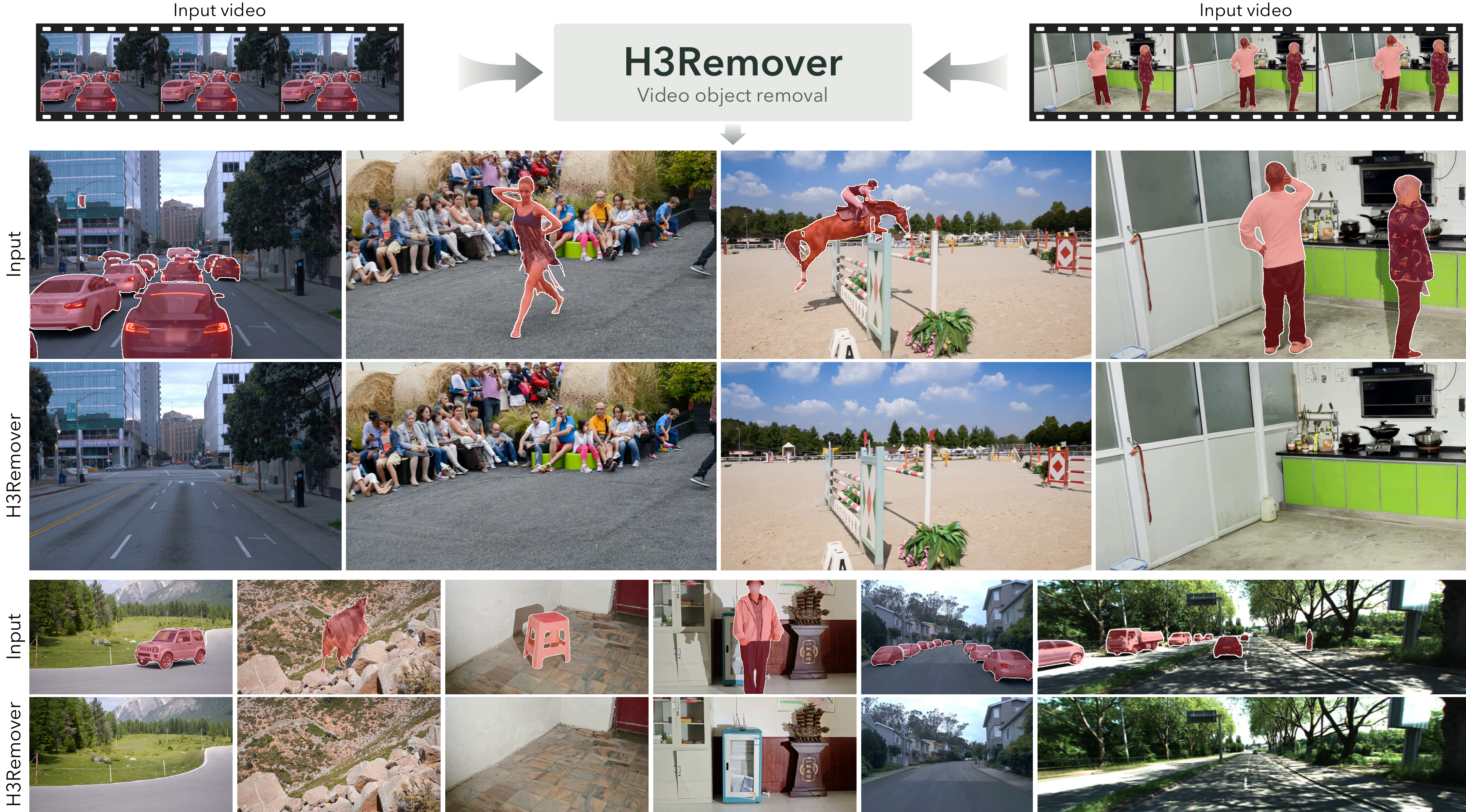}
        \captionof{figure}{
            \textbf{H3Remover enables high-quality video object removal across diverse real-world scenes.}
            Examples span driving, human activities, animals, and indoor scenes,
            showcasing target removal and plausible background completion
            while preserving surrounding structures and non-target content.
            Top filmstrips show input videos.
            Within each paired group, the upper row highlights removal targets
            with translucent masks, and the lower row shows H3Remover results
            at matching timestamps.
        }
        \label{fig:teaser}
        \par\vspace*{10pt}
    \end{minipage}
    \par
}
\makeatother
\maketitle

\begin{abstract}
Pretrained video editors possess rich removal and completion priors, yet may retain targets or alter unrelated content.
We present \textbf{H3Remover}, which reinforces MiniMax-H3 Base Ref2VA for video object removal without paired object-free training targets.
To discourage superficial edits such as recoloring, a target-removal reward favors calibrated depth changes consistent with revealing farther content.
To limit unintended scene changes, a background-geometry reward encourages preservation of relative depth ordering.
To preserve background appearance without retaining unwanted shadows or reflections, an appearance reward constrains RGB changes in source-derived protected regions while exempting targets and estimated effects.
Frozen HPSv3 complements these constraints with perceptual-quality feedback for synthesized content.
Evaluation regions remain fixed across candidates, and DiffusionART optimizes LoRA adapters with the backbone frozen.
Experiments show substantial reconstruction gains over original H3, with qualitative comparisons demonstrating more effective object removal and better preservation of surrounding scene content.
\end{abstract}
\section{Introduction}
\label{sec:introduction}

Video object removal aims to erase unwanted objects and reconstruct the scene behind them while retaining the content that should remain unchanged.
Beyond filling masked pixels, this task requires a careful balance between modifying the target and preserving the surrounding scene.
Flow-guided propagation and spatiotemporal transformers provide effective tools for video inpainting~\cite{liCvpr22vInpainting,zhang2022flow,zhou2023propainter}, while diffusion-based approaches introduce generative priors for recovering missing content~\cite{li2025diffueraserdiffusionmodelvideo,bian2025videopainteranylengthvideoinpainting,zi2025minimax_remover}.
However, the region that needs modification may extend beyond an object's visible silhouette: shadows, reflections, and other associated effects can remain after the object itself disappears~\cite{lu2021,miao2025rose,EffectErase}.
A useful removal model must therefore learn both what to change and what to preserve.

Obtaining paired supervision for this task is challenging: a clean target video must remove both the object and its associated effects while preserving the remaining scene and temporal alignment.
Existing efforts construct such pairs through synthetic rendering or controlled real-world capture~\cite{miao2025rose,EffectErase}.
Rendering depends on curated assets and simulated interactions, while real-world capture requires carefully matched scene conditions, lighting, and camera motion.
These requirements make data construction costly and constrain coverage of diverse real-world scenes and effects.
This motivates adapting a capable pretrained editor without requiring paired object-free target videos.

\begin{figure}[!t]
    \centering
    \includegraphics[width=\columnwidth]
        {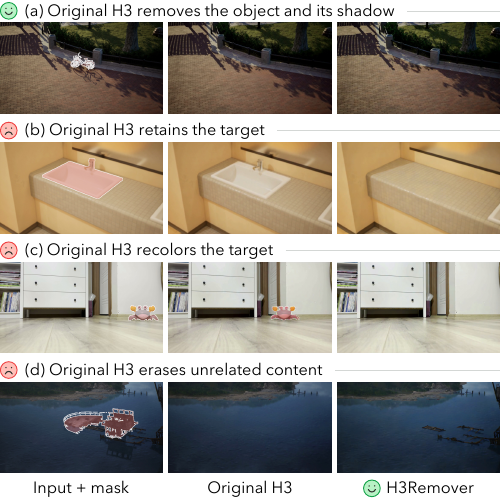}
    \caption{
        \textbf{Motivation: unlocking pretrained removal capabilities.}
        H3 can remove objects and associated effects, including cast
        shadows, but may fail to follow removal instructions or preserve
        unrelated content.
        H3Remover reinforces these existing capabilities while
        constraining changes to the surrounding scene.
        Columns show masked inputs, original H3 outputs, and
        H3Remover outputs; row headings describe original H3's behavior.
    }
    \label{fig:motivation}
\end{figure}

Pretrained video generators and editors offer a promising foundation for such adaptation through their rich visual and editing priors~\cite{yang2024cogvideox,kong2024hunyuanvideo,wan2025,vace}. We build on \textbf{MiniMax-H3 Base Ref2VA} (H3-Base-Ref2VA), the reference-conditioned branch of MiniMax-H3, used here as a video editor.\footnote{\url{https://huggingface.co/MiniMaxAI/MiniMax-H3}}
Our starting observation is that the pretrained model already possesses useful priors for object removal, background completion, and associated-effect editing, but does not consistently express these capabilities under removal instructions.
Figure~\ref{fig:motivation} illustrates this gap.
In Fig.~\ref{fig:motivation}(a), original H3 removes both the bicycle and its cast shadow, demonstrating an existing ability to edit beyond the target silhouette.
In contrast, it retains the target in (b), recolors the toy crab instead of removing it in (c), and erases unrelated scene content along with the target in (d).
These contrasting outcomes motivate strengthening removal instruction following and source-scene preservation: a plausible edited video does not necessarily satisfy the requested edit.

We present \textbf{H3Remover}, a reinforcement learning framework that elicits the editor's removal capabilities while constraining unintended changes.
The desired edit has two coupled requirements: remove the indicated objects and preserve the unrelated scene.
Strict reconstruction outside the target silhouette can preserve unwanted shadows or reflections, whereas rewarding removal alone can tolerate background changes.
General preference models~\cite{xu2023imagereward,Kirstain2023PickaPicAO,liu2025improving} offer quality feedback but do not explicitly encode these source-dependent requirements.
We therefore combine interpretable source-conditioned rewards with auxiliary preference feedback.
Given original and target-highlighted videos and a removal prompt, we optimize LoRA adapters~\cite{hu2022lora} through DiffusionART~\cite{wan2026diffusionartefficientstablescalable}, keeping the backbone frozen and requiring no paired object-free training targets.

Our first two rewards use estimated geometry to distinguish desirable target changes from undesirable background changes.
When an opaque foreground object is removed, the newly revealed surface is typically farther from the camera than the object it replaces.
We use this observation to define a \emph{target-removal reward}: after calibrating source and edited depth predictions to a comparable scale, we reward target pixels whose estimated depth increases.
This signal serves as a proxy for revealing the scene behind the object, rather than a semantic certificate of successful removal.
To constrain the surrounding scene, a complementary \emph{background-geometry reward} penalizes changes in the relative depth ordering of background pixel pairs.
Inspired by the use of depth ranking in SparseNeRF~\cite{wang2022sparsenerf}, this term encourages consistent near--far relationships instead of requiring identical absolute depth predictions.
Both rewards use a frozen Video Depth Anything estimator~\cite{video_depth_anything}, encouraging targeted geometric change while constraining the scene outside the target regions.

Our third reward protects appearance while allowing object-associated effects to change.
We threshold a source-derived blending map from frozen OSOR~\cite{zhou2026osor} into a binary protection mask.
The appearance reward penalizes RGB deviations from the original video only in protected regions, exempting targets and estimated effects.
OSOR's synthesized RGB is never a reconstruction target.
Masks and protection regions are fixed before candidate generation, ensuring that candidates share the same spatial reference.
Frozen HPSv3~\cite{ma2025hpsv3widespectrumhumanpreference} provides complementary visual-quality feedback that encourages plausible reconstruction in regions exposed by object removal, helping reduce black-filled patches and translucent residual artifacts like previous methods.

Our contributions are threefold:
\begin{itemize}
    \item \textbf{Reinforcement of pretrained removal capabilities.}
    We introduce H3Remover, which adapts H3-Base-Ref2VA toward faithful removal instruction following through source-conditioned rewards.
    The framework builds on editing priors and uses DiffusionART-based LoRA optimization without paired object-free target videos during adaptation.

    \item \textbf{Complementary geometry-based rewards.}
    We couple a calibrated target-depth change reward with a background depth-order preservation reward, providing separate feedback for changing the target region and retaining the surrounding scene structure.

    \item \textbf{Effect-aware appearance preservation.}
    We formulate a source-referenced RGB preservation reward with a fixed binary protection mask, allowing changes in estimated effect regions while penalizing unnecessary appearance changes elsewhere.
    Auxiliary preference feedback complements these explicit constraints to guide visual quality.
\end{itemize}
\section{Related Work}
\label{sec:related_work}

\paragraph{Video generation and editing.}
Diffusion models~\cite{ho2020denoising,song2020denoising}, latent diffusion~\cite{rombach2021highresolution}, and diffusion transformers~\cite{Peebles2022DiT} support scalable visual generation.
CogVideoX~\cite{yang2024cogvideox}, HunyuanVideo~\cite{kong2024hunyuanvideo}, Wan~\cite{wan2025}, and VACE~\cite{vace} extend generative priors and conditioning mechanisms to video.
Following image editing approaches such as Prompt-to-Prompt~\cite{hertz2022prompt} and InstructPix2Pix~\cite{brooks2022instructpix2pix}, video editors address correspondence through adaptation~\cite{wu2023tune}, cross-frame attention~\cite{text2video-zero}, attention control~\cite{qi2023fatezero,liu2023videop2p}, and feature propagation~\cite{tokenflow2023}.
Pix2Video~\cite{ceylan2023pix2video}, Rerender A Video~\cite{yang2023rerender}, and AnyV2V~\cite{ku2024anyv2v} transfer image-level edits to video; CoDeF~\cite{ouyang2023codef} separates canonical content and temporal deformation.
H3Remover studies task-specific adaptation feedback for an existing video editor.

\paragraph{Video removal and object-associated effects.}
Video inpainting uses transformer fusion~\cite{Liu_2021_FuseFormer}, flow-guided synthesis~\cite{liCvpr22vInpainting,zhang2022flow}, and dual-domain propagation~\cite{zhou2023propainter}.
DiffuEraser~\cite{li2025diffueraserdiffusionmodelvideo}, VideoPainter~\cite{bian2025videopainteranylengthvideoinpainting}, and MiniMax-Remover~\cite{zi2025minimax_remover} exploit diffusion priors for completion and removal.
Omnimatte~\cite{lu2021}, OmnimatteRF~\cite{Lin_2023_ICCV}, Generative Omnimatte~\cite{gen-omnimatte}, and OmnimatteZero~\cite{samuel2025omnimattezero} associate objects with their effects through decomposition and generative priors.
ROSE~\cite{miao2025rose}, EffectErase~\cite{EffectErase}, Object-WIPER~\cite{kushwaha2026objectwipertrainingfreeobject}, EffectLearner~\cite{wu2026effectlearnerworldawareobjecteffectreasoning}, and VOID~\cite{motamed2026void} address removal with associated effects or interactions.
Image methods include ObjectClear~\cite{zhao2026objectclear} and OSOR~\cite{zhou2026osor}.
Our focus is adapting an editor using source-referenced rewards.
We reuse OSOR to locate regions where appearance preservation should be relaxed; its output is neither a clean RGB target nor an effect-removal success score.
Thus, effect localization is supplied by an expert, while our design specifies how it enters the adaptation objective.

\paragraph{Geometry and reward-guided adaptation.}
Depth Anything~\cite{depthanything}, Depth Anything V2~\cite{depth_anything_v2}, Marigold~\cite{ke2023repurposing}, and Video Depth Anything~\cite{video_depth_anything} provide depth priors.
SparseNeRF~\cite{wang2022sparsenerf} uses depth ranking in novel-view synthesis; we use ordering to compare a source video with its edited counterpart.
DDPO~\cite{black2023ddpo}, AlignProp~\cite{prabhudesai2023aligning}, and Diffusion-DPO~\cite{wallace2023diffusion} study reward or preference optimization.
ImageReward~\cite{xu2023imagereward}, PickScore~\cite{Kirstain2023PickaPicAO}, VideoReward~\cite{liu2025improving}, and HPSv3~\cite{ma2025hpsv3widespectrumhumanpreference} supply quality or preference signals.
Flow-GRPO~\cite{liu2025flow}, DanceGRPO~\cite{xue2025dancegrpo}, DiffusionNFT~\cite{zheng2025diffusionnft}, and DiffusionART~\cite{wan2026diffusionartefficientstablescalable} develop generative-model optimization strategies.
We use DiffusionART with LoRA~\cite{hu2022lora} to optimize the editor with these task-specific signals.
Our focus is the regional reward construction: encourage target change, constrain background ordering, and protect appearance outside estimated effect regions.
Frozen HPSv3 complements these regional rewards with visual-quality feedback.
\section{H3Remover}
\label{sec:method}

\begin{figure*}[!t]
    \centering
    \includegraphics[width=\textwidth]
        {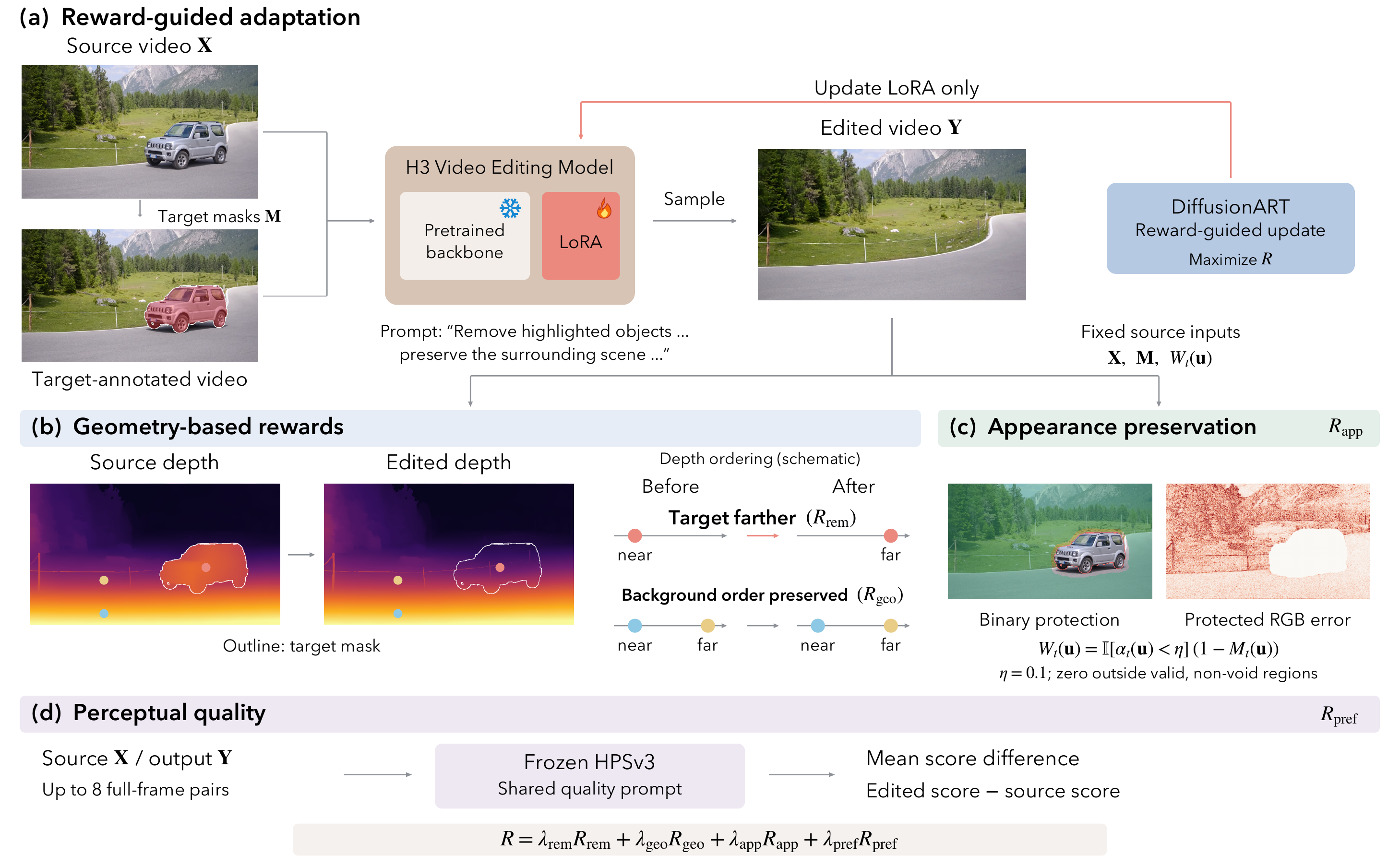}
    \caption{
        \textbf{Overview of H3Remover.}
        (a) Source and target-annotated videos condition the frozen
        editor with trainable LoRA adapters.
        (b) Geometry rewards encourage farther estimated target depth and
        preservation of background depth ordering.
        (c) Binary OSOR-derived protection constrains RGB changes
        outside targets and estimated effects.
        (d) Frozen HPSv3 supplies preference feedback.
        The combined reward guides DiffusionART updates.
        Auxiliary maps illustrate computation; depth-relation
        diagrams are schematic.
    }
    \label{fig:pipeline}
\end{figure*}

H3Remover adapts a pretrained H3-Base-Ref2VA editor using source-referenced rewards rather than paired object-free targets (Fig.~\ref{fig:pipeline}).
The design separates the desired target change from constraints on the surrounding scene; all auxiliary estimators and the backbone remain frozen.

\subsection{Source-Conditioned Editing}
\label{sec:conditioning}

Let $\mathbf{X}=\{\mathbf{X}_t\}_{t=1}^{N}$
denote a source video with RGB values in $[0,1]$,
and let $\mathbf{Y}$ denote a candidate edit.
For frame $t$, $\mathcal{I}_t$ indexes visible target instances,
$\Omega_t^j$ is the pixel support of instance $j$,
and $\mathbf{M}_t$ is their binary union mask.
The editor receives the original video,
a synchronized copy with translucent target masks
and white contours, and a removal instruction.
The annotated copy identifies targets; the original supplies unobscured appearance.
The prompt requests removal of targets and associated effects while preserving unrelated content and camera motion.

Training targets use dataset annotations
and cached SAM~3.1 tracks~\cite{carion2025sam3segmentconcepts}
for the driving data.
Target identities, masks, valid regions, and source-frame eligibility
are fixed before sampling and shared across candidates.
Edits cannot redefine their evaluation regions.
Padding, void regions, and unreliable source frames provide no removal credit.

\subsection{Geometry-Based Rewards}
\label{sec:geometry_rewards}

\paragraph{Target removal.}
An RGB change inside a target mask can indicate recoloring
rather than deletion.
We instead exploit the expectation that removing an occluding
foreground object reveals farther content.
A frozen Video Depth Anything estimator~\cite{video_depth_anything}
predicts source and edited inverse depths,
$\mathbf{D}_t^{\mathrm{x}}$ and $\mathbf{D}_t^{\mathrm{y}}$,
with larger values indicating nearer surfaces.
Because independent predictions can differ in scale and shift,
we robustly fit one positive affine mapping over the clip,
$\widehat{\mathbf{D}}_t^{\mathrm{y}}
=a\mathbf{D}_t^{\mathrm{y}}+b\mathbf{1}$,
using background anchors outside source target bounding boxes.
Disjoint anchor subsets provide fitting and residual estimation.
Calibration assumes enough comparable background anchors remain;
it uses depth alone, without RGB filtering or spatial alignment.
Defining
$\delta_t(\mathbf{u})=
D_t^{\mathrm{x}}(\mathbf{u})
-\widehat D_t^{\mathrm{y}}(\mathbf{u})$,
we reward the fraction of target pixels exhibiting
a sufficiently large decrease in inverse depth:
\begin{equation}
    r_t^{\mathrm{rem}}=
    \frac{1}{|\mathcal{I}_t|}
    \sum_{j\in\mathcal{I}_t}
    \frac{1}{|\Omega_t^j|}
    \sum_{\mathbf{u}\in\Omega_t^j}
    \mathbb{I}[\delta_t(\mathbf{u})>\tau].
    \label{eq:target_reward}
\end{equation}
Here $\mathbb{I}$ is an indicator and
$\tau=\max(0.01s_{\mathrm{x}},e_{95})$,
where $s_{\mathrm{x}}$ is the source fitting-anchor
$95$th--$5$th percentile depth span
and $e_{95}$ is the held-out $95$th-percentile
absolute calibration residual.
The threshold discounts depth changes within the calibration residual.
A fraction score limits the benefit of excessive depth change,
and instance averaging prevents large targets from dominating.
This proxy relies on the foreground-occluder assumption rather than testing semantic absence. In many scenes, particularly road scenes, removing foreground vehicles or pedestrians reveals farther surfaces; reinforcement on such cases may also benefit removal in scenes where this depth assumption does not hold through shared editing capabilities.

\paragraph{Background geometry.}
Farther target pixels can coexist with background rearrangement.
We constrain the background while avoiding sensitivity to
numerical depth scale.
Inspired by depth-ranking constraints~\cite{wang2022sparsenerf},
we instead compare relative ordering at corresponding
background coordinates.
A source-fixed pair set $\mathcal{P}_t$ uses horizontal
and vertical offsets of $1$, $8$, $32$, and $128$ pixels,
with both endpoints outside $\mathbf{M}_t$.
For each depth map,
$o_t(\mathbf{u},\mathbf{v})\in\{-1,0,+1\}$
describes near, tied, or far ordering;
differences within $1\%$ of that map's background
$95$th--$5$th percentile span are treated as ties.
The reward is
\begin{equation}
    r_t^{\mathrm{geo}}=
    -\frac{1}{|\mathcal{P}_t|}
    \sum_{(\mathbf{u},\mathbf{v})\in\mathcal{P}_t}
    \mathbb{I}\!\left[
        o_t^{\mathrm{x}}(\mathbf{u},\mathbf{v})
        \neq
        o_t^{\mathrm{y}}(\mathbf{u},\mathbf{v})
    \right].
    \label{eq:geometry_reward}
\end{equation}
Relative thresholds provide invariance to positive affine depth
transformations; ties reduce sensitivity on nearly flat surfaces.
Source-valid pairs with invalid edited depths count
as mismatches rather than disappearing from the denominator.
This discourages changes in background ordering without
aligning away spatial drift.
A separate RGB constraint complements this structural feedback.

\subsection{Effect-Aware Appearance Preservation}
\label{sec:appearance_reward}

Depth ordering leaves color and texture underconstrained,
but RGB preservation outside the silhouette can retain unwanted shadows
or reflections.
We therefore separate protected background from editable effects.
A frozen OSOR model~\cite{zhou2026osor} supplies
a source-derived blending map $\boldsymbol{\alpha}_t$.
Within the valid, non-void image region,
we define binary protection as
$W_t(\mathbf{u})=
\mathbb{I}[\alpha_t(\mathbf{u})<\eta]
\bigl(1-M_t(\mathbf{u})\bigr)$,
with $\eta=0.1$, and set it to zero elsewhere.
The RGB reward is
\begin{equation}
    r_t^{\mathrm{app}}=-
    \frac{
        \sum_{\mathbf{u}}W_t(\mathbf{u})
        \|\mathbf{Y}_t(\mathbf{u})
        -\mathbf{X}_t(\mathbf{u})\|_1
    }{
        3\sum_{\mathbf{u}}W_t(\mathbf{u})
    }.
    \label{eq:appearance_reward}
\end{equation}
Binary exclusion removes RGB preservation pressure inside estimated
effect regions; preservation elsewhere remains a soft penalty.
Area normalization makes errors comparable across protected supports.
The source-fixed map prevents candidates from expanding editable regions.
OSOR supplies an editable-region estimate, not an effect-removal score;
its synthesized RGB is never a reconstruction target.

\subsection{Auxiliary Perceptual Preference}
\label{sec:preference_reward}

To assess synthesized appearance beyond geometry and protected pixels,
we add a frozen HPSv3 scorer~\cite{ma2025hpsv3widespectrumhumanpreference}.
Let $\mathcal{S}$ contain up to eight uniformly spaced
source-eligible timestamps, fixed before candidate generation.
At each timestamp, $h(\mathbf{I},q)$ denotes the scorer's
mean-preference output for the full valid image region
under a fixed photographic-quality prompt $q$.
We compute
\begin{equation}
    R_{\mathrm{pref}}=
    \frac{1}{|\mathcal{S}|}
    \sum_{t\in\mathcal{S}}
    \left[
        h(\mathbf{Y}_t,q)-h(\mathbf{X}_t,q)
    \right].
    \label{eq:preference_reward}
\end{equation}
The shared prompt emphasizes natural lighting, coherent geometry,
intact textures, and freedom from artifacts.
Frames are scored independently without target crops or mask overlays.
Subtracting the source score supplies a constant within each candidate
group and does not change its ranking.
HPSv3 assesses image quality, not requested target absence.

\subsection{Reward-Guided LoRA Adaptation}
\label{sec:optimization}

For $k\in\{\mathrm{rem},\mathrm{geo},\mathrm{app}\}$,
we aggregate eligible frame rewards as
$R_k=N^{-1}\sum_{t\in\mathcal{T}}r_t^k$,
where $\mathcal{T}$ is the source-fixed eligible-frame set.
Normalization by the full clip length prevents a few eligible
frames from receiving the same score as full temporal coverage.
The total reward is
\begin{equation}
\begin{split}
    R={}&
    \lambda_{\mathrm{rem}}R_{\mathrm{rem}}
    +\lambda_{\mathrm{geo}}R_{\mathrm{geo}}
    +\lambda_{\mathrm{app}}R_{\mathrm{app}}\\
    &+\lambda_{\mathrm{pref}}R_{\mathrm{pref}},
\end{split}
    \label{eq:total_reward}
\end{equation}
with coefficients $(1,1,5,0.05)$ for the full model.
Both preservation terms are negative errors,
so larger $R$ is better.
Insufficient required support, failed depth calibration,
or failed preference scoring makes a candidate unavailable;
incomplete four-candidate groups are excluded
rather than assigning failures zero reward.

We maximize expected reward using
DiffusionART~\cite{wan2026diffusionartefficientstablescalable},
updating only LoRA parameters
$\boldsymbol{\phi}$~\cite{hu2022lora}.
Four candidates per condition provide a reward--endpoint
covariance direction, which is normalized and held fixed
during replay-based velocity updates.
The update accommodates discrete rewards without differentiating through the frozen estimators; only generated video tokens enter the loss.
Inference uses the source, target annotation, and prompt without
reward evaluation; reward estimators add adaptation-time cost only.

Temporal coherence relies on the video editor's prior and framewise
source constraints; the reward contains no explicit temporal term.
\section{Experiments}
\label{sec:experiments}


\begin{figure*}[!t]
    \centering
    \includegraphics[width=\textwidth]
        {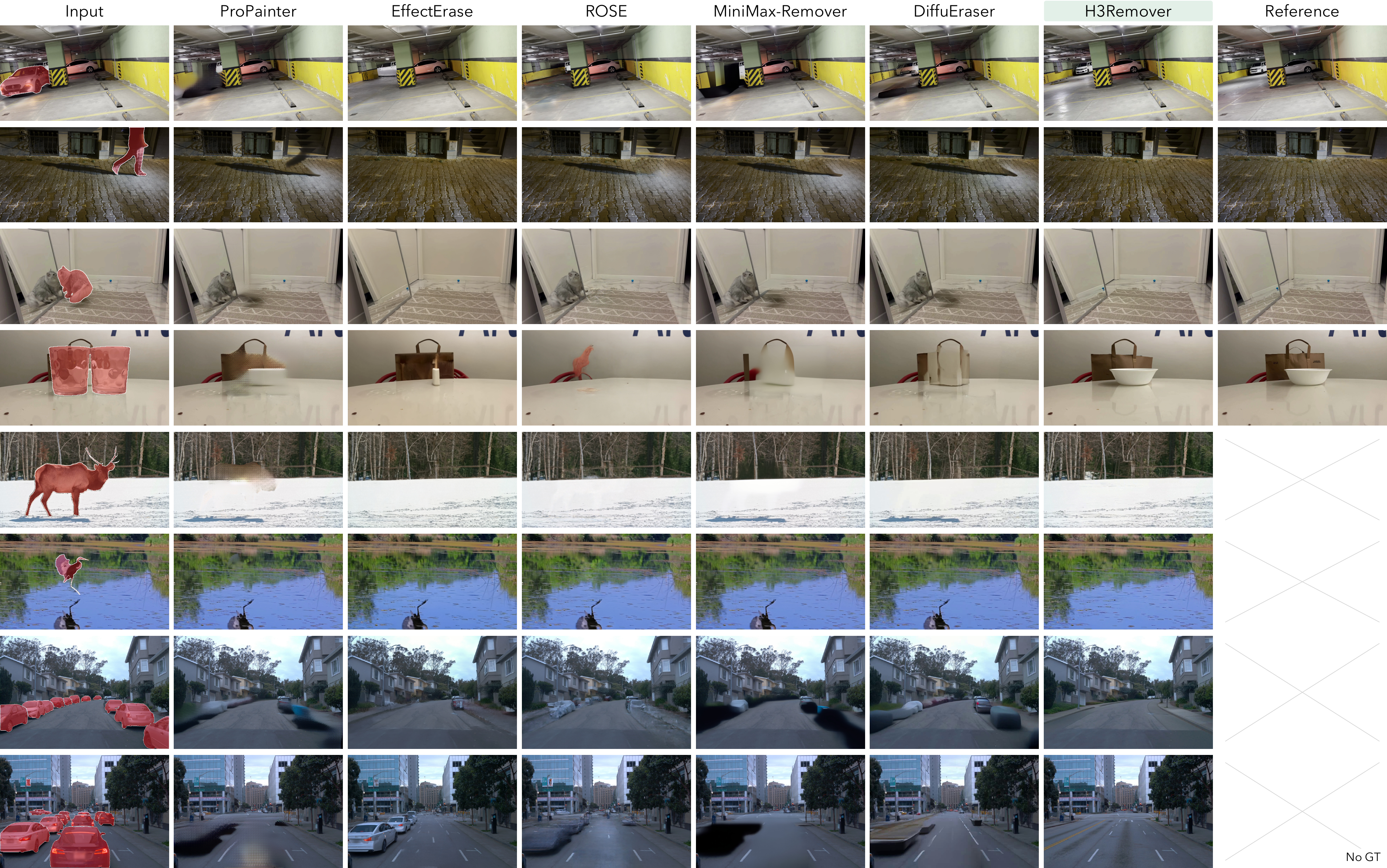}
    \caption{
        \textbf{Qualitative comparison across diverse removal scenarios.}
        Each row uses matched inputs, timestamps, and crops;
        translucent masks mark targets.
        The bottom four cases have no clean reference (gray crosses).
        In the bottom two driving rows, H3Remover is prompted to remove
        all visible vehicles and pedestrians, whereas competing methods
        use the supplied target masks.
    }
    \label{fig:method_comparison}
\end{figure*}

\begin{table*}[!t]
    \centering
    \caption{
        \textbf{Quantitative comparison on PROVE-M.}
        Methods use identical valid video windows.
        Original H3 is the pretrained reference.
        $S/F/U$ denotes automated removal success/failure/unknown;
        the $S/N$ denominator includes unknowns.
        Automated judgments are exploratory due to display-layout differences.
        The top three results are highlighted in
        \protect\colorbox{rankfirst}{first},
        \protect\colorbox{ranksecond}{second}, and
        \protect\colorbox{rankthird}{third}, respectively.
    }
    \label{tab:paired_comparison}

    \begingroup
    \small
    \setlength{\tabcolsep}{4pt}
    \renewcommand{\arraystretch}{1.06}

    \begin{tabular*}{\textwidth}{@{\extracolsep{\fill}}lcccccccc}
        \toprule
        Method
        & PSNR$\uparrow$
        & SSIM$\uparrow$
        & LPIPS$\downarrow$
        & RC-S$\uparrow$
        & RC-T$\downarrow$
        & FVD$\downarrow$
        & $S/N$ (\%)$\uparrow$
        & S/F/U \\
        \midrule

        DiffuEraser~\cite{li2025diffueraserdiffusionmodelvideo}
        & 22.277 & 0.8403 & 0.19516
        & \cellcolor{rankthird}0.4807
        & 0.5416 & 795.03 & 83.75 & 67/12/1 \\

        ProPainter~\cite{zhou2023propainter}
        & 22.724 & 0.8586 & 0.18210
        & 0.4280 & 0.6335 & 769.89 & 82.50 & 66/13/1 \\

        MiniMax-Remover~\cite{zi2025minimax_remover}
        & 22.334 & 0.8584 & 0.17714
        & 0.4755
        & \cellcolor{rankthird}0.4368
        & 728.90 & 77.50 & 62/16/2 \\

        ROSE~\cite{miao2025rose}
        & \cellcolor{ranksecond}27.024
        & \cellcolor{ranksecond}0.8965
        & \cellcolor{ranksecond}0.10492
        & 0.4773 & 0.6540
        & \cellcolor{rankthird}373.37
        & \cellcolor{rankthird}88.75
        & 71/6/3 \\

        EffectErase~\cite{EffectErase}
        & \cellcolor{rankthird}26.085
        & \cellcolor{rankthird}0.8917
        & \cellcolor{rankthird}0.10708
        & \cellcolor{ranksecond}0.5120
        & \cellcolor{ranksecond}0.2953
        & \cellcolor{ranksecond}193.35
        & \cellcolor{ranksecond}90.00
        & 72/6/2 \\

        \midrule[0.4pt]

        Original H3
        & 21.268 & 0.7718 & 0.21446
        & 0.3694 & 0.4719 & 732.74
        & 50.00 & 40/39/1 \\

        H3Remover
        & \cellcolor{rankfirst}28.419
        & \cellcolor{rankfirst}0.9003
        & \cellcolor{rankfirst}0.08506
        & \cellcolor{rankfirst}0.5204
        & \cellcolor{rankfirst}0.2275
        & \cellcolor{rankfirst}112.21
        & \cellcolor{rankfirst}91.25
        & 73/6/1 \\

        \bottomrule
    \end{tabular*}
    \endgroup
\end{table*}

\begin{table*}[!t]
    \centering
    \caption{
        \textbf{Exploratory results without clean references.}
        Ranks are numerical; judgment layouts and driving
        target scopes differ across methods, as detailed in the text.
        $S/N$ includes unknowns.
    }
    \label{tab:no_reference_main}

    \begingroup
    \small
    \setlength{\tabcolsep}{3pt}
    \renewcommand{\arraystretch}{1.08}

    \begin{tabular*}{\textwidth}{@{\extracolsep{\fill}}lccccccccc}
        \toprule
        & \multicolumn{3}{c}{WIPER}
        & \multicolumn{3}{c}{KITTI-360}
        & \multicolumn{3}{c}{Waymo} \\
        \cmidrule(lr){2-4}
        \cmidrule(lr){5-7}
        \cmidrule(lr){8-10}

        Method
        & RC-S$\uparrow$
        & RC-T$\downarrow$
        & $S/N$ (\%)$\uparrow$
        & RC-S$\uparrow$
        & RC-T$\downarrow$
        & $S/N$ (\%)$\uparrow$
        & RC-S$\uparrow$
        & RC-T$\downarrow$
        & $S/N$ (\%)$\uparrow$ \\
        \midrule

        ProPainter~\cite{zhou2023propainter}
        & 0.2795 & 0.5495 & 63.33
        & 0.3252
        & \cellcolor{rankthird}1.5348
        & \cellcolor{rankthird}80.00
        & 0.3471 & 0.7039
        & \cellcolor{ranksecond}80.00 \\

        DiffuEraser~\cite{li2025diffueraserdiffusionmodelvideo}
        & 0.4252 & 0.4313 & 76.67
        & \cellcolor{rankthird}0.3910
        & 1.5895 & 73.33
        & 0.3878 & 0.7014 & 60.00 \\

        ROSE~\cite{miao2025rose}
        & \cellcolor{rankthird}0.4408
        & 0.4153
        & \cellcolor{rankthird}90.00
        & 0.3710 & 1.6273
        & \cellcolor{ranksecond}86.67
        & 0.3732 & 0.8149
        & \cellcolor{rankthird}66.67 \\

        MiniMax-Remover~\cite{zi2025minimax_remover}
        & 0.4205 & 0.2817 & 78.33
        & \cellcolor{ranksecond}0.4040
        & 1.5463
        & \cellcolor{rankthird}80.00
        & \cellcolor{ranksecond}0.4036
        & 0.7099
        & \cellcolor{ranksecond}80.00 \\

        EffectErase~\cite{EffectErase}
        & \cellcolor{rankfirst}0.5134
        & \cellcolor{ranksecond}0.1963
        & \cellcolor{ranksecond}91.67
        & 0.3775
        & \cellcolor{ranksecond}1.4727
        & \cellcolor{rankthird}80.00
        & \cellcolor{rankthird}0.3905
        & \cellcolor{rankthird}0.6402
        & \cellcolor{rankthird}66.67 \\

        \midrule[0.4pt]

        Original H3
        & 0.4294
        & \cellcolor{rankthird}0.2021
        & 71.67
        & 0.2514 & 1.7131 & 6.67
        & 0.3552
        & \cellcolor{ranksecond}0.5607
        & \cellcolor{rankthird}66.67 \\

        H3Remover
        & \cellcolor{ranksecond}0.5001
        & \cellcolor{rankfirst}0.1413
        & \cellcolor{rankfirst}100.00
        & \cellcolor{rankfirst}0.4130
        & \cellcolor{rankfirst}0.8810
        & \cellcolor{rankfirst}93.33
        & \cellcolor{rankfirst}0.4322
        & \cellcolor{rankfirst}0.4176
        & \cellcolor{rankfirst}93.33 \\

        \bottomrule
    \end{tabular*}
    \endgroup
\end{table*}

\begin{figure*}[!t]
    \centering

    \begin{minipage}{\textwidth}
        \centering
        \captionof{table}{
            \textbf{Exploratory reward-configuration comparison on 80 cases:}
            32 PROVE-M, 24 ROSE, and 24 BeyondMasks tasks selected by
            reproducible stratified sampling.
            Checkmarks indicate enabled rewards;
            \protect\colorbox{rankfirst}{green} marks the best value.
        }
        \label{tab:reward_ablation}

        \begingroup
        \def\AblationTableWidth{0.90\textwidth}

        \small
        \renewcommand{\arraystretch}{1.08}
        \setlength{\tabcolsep}{4pt}

        \begin{tabularx}{\AblationTableWidth}{
            l *{9}{>{\centering\arraybackslash}X}
        }
            \toprule
            Setting
            & $R_{\mathrm{rem}}$
            & $R_{\mathrm{geo}}$
            & $R_{\mathrm{app}}$
            & $R_{\mathrm{pref}}$
            & PSNR$\uparrow$
            & SSIM$\uparrow$
            & LPIPS$\downarrow$
            & RC-S$\uparrow$
            & RC-T$\downarrow$ \\
            \midrule

            Original H3
            & -- & -- & -- & --
            & 22.390 & 0.7886 & 0.18573 & 0.3617 & 0.3175 \\

            \midrule

            \multirow{4}{*}{H3Remover}
            & $\checkmark$ & & &
            & 24.574 & 0.8348 & 0.12273
            & \cellcolor{rankfirst}0.4995
            & 0.2339 \\

            & $\checkmark$ & $\checkmark$ & &
            & 26.741 & 0.8768 & 0.09593 & 0.4838 & 0.2069 \\

            & $\checkmark$ & $\checkmark$ & $\checkmark$ &
            & 27.464 & 0.8867 & 0.10683 & 0.4767 & 0.2218 \\

            & $\checkmark$ & $\checkmark$ & $\checkmark$ & $\checkmark$
            & \cellcolor{rankfirst}27.869
            & \cellcolor{rankfirst}0.8899
            & \cellcolor{rankfirst}0.09501
            & 0.4933
            & \cellcolor{rankfirst}0.1969 \\

            \bottomrule
        \end{tabularx}
        \endgroup
    \end{minipage}

    \par\vspace{8pt}

    \begin{minipage}{\textwidth}
        \centering
        \includegraphics[width=\textwidth]
            {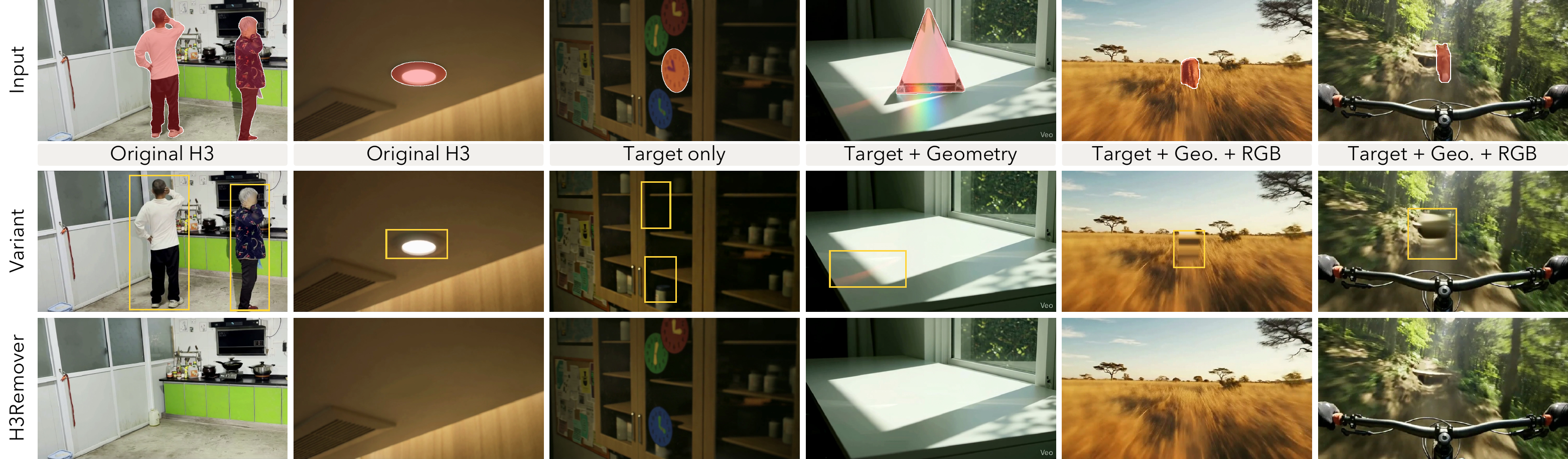}
        \captionof{figure}{
            \textbf{Qualitative reward-configuration comparisons.}
            Top: masked inputs.
            Middle: indicated configurations, with yellow boxes
            highlighting retained targets, unintended deletion,
            refractive residue, or local blur.
            Bottom: full-model results at matching timestamps and crops.
            Selected examples illustrate differences between complete
            configurations.
        }
        \label{fig:reward_ablation}
    \end{minipage}
\end{figure*}

\begin{figure*}[!t]
    \centering
    \includegraphics[width=\textwidth]
        {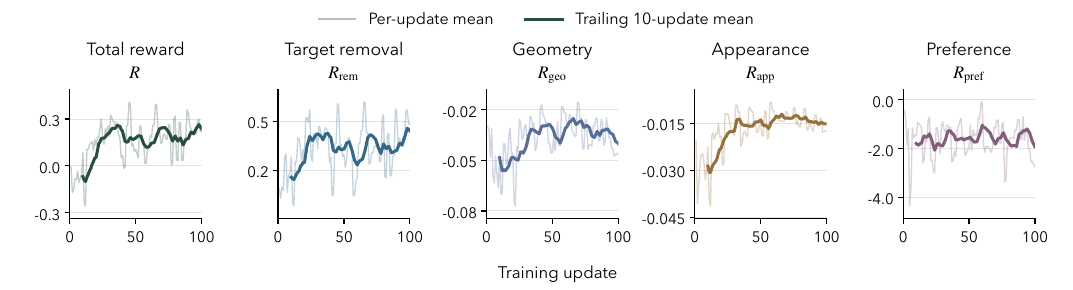}
    \caption{
        \textbf{Training reward dynamics of the full model.}
        Curves show available-candidate means over the training
        updates; light lines are raw values and dark lines are trailing
        10-update averages.
        Components use the signed, unweighted definitions in the Method;
        the total uses weights $(1,1,5,0.05)$.
        Training batches and valid-candidate counts vary across updates.
    }
    \label{fig:training_rewards}
\end{figure*}

\subsection{Implementation Details}
\label{sec:implementation}

\paragraph{Adaptation.}
We adapt H3-Base-Ref2VA on 970 removal tasks: 290 ROSE, 214 Waymo, 214 KITTI-360, 195 MOSE, and 57 DAVIS2017.
Only source videos and target annotations enter adaptation; object-free references are not training targets.
Rank-16 LoRA adapters update attention projections, with the backbone and reward estimators frozen.
AdamW uses a peak learning rate of $5\times10^{-4}$, 20 warm-up updates, and cosine decay.
We sample four candidates per condition, reporting the 100-update checkpoint.
Rollouts and evaluation use 10 and 50 sampling points, respectively.
The full model uses weights $(\lambda_{\mathrm{rem}},\lambda_{\mathrm{geo}},\lambda_{\mathrm{app}},\lambda_{\mathrm{pref}})=(1,1,5,0.05)$ and protection threshold $\eta=0.1$.

\paragraph{Evaluation datasets.}
We evaluate H3Remover on PROVE-M~\cite{li2026prove}, WIPER~\cite{kushwaha2026objectwipertrainingfreeobject}, KITTI-360~\cite{Liao2022PAMI}, and Waymo~\cite{Sun_2020_CVPR}.
PROVE-M contains 80 paired tasks with object-free references, enabling reference-based evaluation of removal and scene reconstruction.
WIPER contributes 60 tasks without clean references, complementing the paired evaluation.
KITTI-360 and Waymo each contribute 15 driving clips, covering vehicle and pedestrian removal in road scenes.
The 30 driving clips originate from 10 independent scenes.
For datasets without clean references, we report RC-S, RC-T, and automated target-disappearance judgments.

\paragraph{Evaluation protocol.}
Methods are evaluated on a shared 124-frame, 24-Hz window and a $544\times960$ canvas; spatial padding and repeated tail frames are excluded.
Baselines retain method-specific inference settings, so our common-window results are not reproductions of their published native-resolution scores.
We report PSNR, SSIM, AlexNet-based LPIPS, RC-S, RC-T, and FVD, using a fixed I3D-based protocol for FVD.
Without clean references, RC-S ($\uparrow$) measures local feature coherence between restored regions and surrounding background, while RC-T ($\downarrow$) measures temporal discrepancy in shared restored regions across adjacent frames~\cite{li2026prove}.
GPT-6-sol judges target disappearance from up to eight matched frames, with method identities anonymized.
We report $S/N$, where $N=S+F+U$ includes success, failure, and unknown judgments.
This automated evaluation checks target disappearance in sampled frames.

\subsection{Comparison with Existing Methods}
\label{sec:comparison}

\paragraph{Paired reconstruction.}
Table~\ref{tab:paired_comparison} compares H3Remover with ProPainter~\cite{zhou2023propainter}, DiffuEraser~\cite{li2025diffueraserdiffusionmodelvideo}, MiniMax-Remover~\cite{zi2025minimax_remover}, ROSE~\cite{miao2025rose}, and EffectErase~\cite{EffectErase} on PROVE-M.
H3Remover leads all six image/video metrics.
Against original H3, PSNR improves by 7.151 dB, LPIPS and FVD decrease by 60.3\% and 84.7\%, and sampled-frame confirmed removals increase from 40/80 to 73/80.
Against the best baseline per metric, PSNR improves by 1.395 dB, LPIPS decreases by 18.9\%, and FVD decreases by 42.0\%.
Confirmed removal is comparable to EffectErase (73/80 versus 72/80), with better reconstruction quality.

\paragraph{Exploratory unpaired evaluation.}
Table~\ref{tab:no_reference_main} covers 60 WIPER clips and 30 driving clips from ten scenes, equally split between KITTI-360 and Waymo.
H3Remover improves both RC metrics over original H3 on all three sets; sampled-frame confirmed removals rise from 43/60 to 60/60, 1/15 to 14/15, and 10/15 to 14/15, respectively.
It also records the lowest RC-T and highest confirmed-removal fraction on all three sets, and the highest RC-S on both driving sets.
These comparisons remain exploratory: judgment layouts differ, and driving prompts target all visible vehicles/people for both H3 variants but supplied-mask targets for competing methods.
Unknowns remain in the denominator; source videos are not clean ground truth.

\paragraph{Visual evidence.}
Figure~\ref{fig:method_comparison} shows scenarios involving target remnants, shadows, reflections, and non-target structure.
The selected examples highlight object removal and background preservation behavior; the driving rows use the target scopes described above.
Compared with competing methods, H3Remover leaves fewer visible target remnants and associated effects in the illustrated cases.
The revealed regions also blend more naturally with adjacent content, with fewer conspicuous artifacts and better continuity of background textures.
Meanwhile, surrounding structures and non-target objects are better retained, illustrating a favorable balance between removal completeness and scene preservation.

\subsection{Reward Analysis}
\label{sec:ablation}

\paragraph{Adapted versus pretrained editor.}
Table~\ref{tab:reward_ablation} compares original H3 with four adapted configurations using identical source videos, masks, prompts, and initial sampling noise at evaluation.
The full configuration improves PSNR from 22.390 to 27.869 dB and reduces LPIPS by 48.8\%.
These gains demonstrate improved reconstruction quality over the pretrained editor under matched evaluation inputs.

\paragraph{Configuration trade-offs.}
Adding geometry to target-only training improves PSNR by 2.167 dB.
Binary RGB protection further improves PSNR and SSIM, with trade-offs in LPIPS and RC-T.
The full configuration achieves the best PSNR, SSIM, LPIPS, and RC-T, while target-only training retains the highest RC-S.
Figure~\ref{fig:reward_ablation} illustrates complementary reward roles: target feedback encourages removal of objects retained by original H3; geometry constrains unintended deletion of non-target clocks; protected RGB feedback supports cleaner tabletop appearance with reduced refractive residue; and HPSv3 helps reduce local blur and translucent artifacts.
These examples compare complete configurations rather than isolate individual reward effects.

\paragraph{Optimization dynamics.}
Figure~\ref{fig:training_rewards} summarizes training statistics.
Between updates 1--10 and 91--100, mean total reward rises from $-0.069$ to $0.238$ and target-depth reward from $0.213$ to $0.452$; protected RGB error falls from $0.0286$ to $0.0151$ and background-order error from $0.0487$ to $0.0417$.
These trends indicate stronger target-depth responses alongside improved appearance and geometric preservation.
HPSv3 feedback fluctuates without an upward trend; its intended role is a perceptual constraint that regularizes visual quality alongside task-specific rewards.
\section{Conclusion}
\label{sec:conclusion}

We presented H3Remover, a reinforcement learning framework that unlocks pretrained video editors' object-removal capabilities without paired object-free targets during adaptation.
By combining geometry-based constraints, protected appearance preservation, and perceptual quality feedback, H3Remover encourages faithful removal while limiting unintended changes to the surrounding scene.
Experiments demonstrate clear improvements over the original H3 model, with effective removal of objects and their associated effects, plausible background reconstruction, and improved scene preservation.
Our findings highlight source-conditioned reward design as a practical approach to strengthening existing video editing capabilities without paired supervision.



{
    \small
    \bibliographystyle{ieeenat_fullname}
    \bibliography{main}
}


\end{document}